\documentclass[11pt]{article}
\usepackage{amsmath}

\usepackage[preprint]{acl}

\usepackage{times}
\usepackage{latexsym}

\usepackage[T1]{fontenc}

\usepackage[utf8]{inputenc}

\usepackage{microtype}

\usepackage{inconsolata}

\usepackage{graphicx}
\usepackage{amsfonts}
\usepackage{booktabs}
\usepackage[table]{xcolor}
\usepackage{array}
\usepackage{subcaption}
\usepackage{tabularx}
\usepackage{graphicx}

\definecolor{myblue1}{RGB}{31, 119, 180}
\definecolor{myblue2}{RGB}{23, 190, 207}
\definecolor{myred1}{RGB}{214, 39, 40}

\usepackage{titlesec}
\newcommand{\compacttitlespacing}{%
  \titlespacing*{\section}{0pt}{7pt plus 1pt minus 1pt}{4pt plus 1pt minus 1pt}
  \titlespacing*{\subsection}{0pt}{2pt plus 0pt minus 0pt}{2pt plus 0pt minus 0pt}
  \titlespacing*{\subsubsection}{0pt}{4pt plus 1pt minus 1pt}{2pt plus 1pt minus 1pt}
}

\newcommand{\appendixtitlespacing}{%
  \titlespacing*{\section}{0pt}{3.5ex plus 1ex minus .2ex}{2.3ex plus .2ex}
  \titlespacing*{\subsection}{0pt}{3.25ex plus 1ex minus .2ex}{1.5ex plus .2ex}
  \titlespacing*{\subsubsection}{0pt}{3.25ex plus 1ex minus .2ex}{1.5ex plus .2ex}
}

\title{Slots, Transitions, Loops: Learning Composable World Models for ARC}

\author{Gege Gao \\
  University of Tübingen \\
  \texttt{gege.gao@uni-tuebingen.de} 
  \\\And
  Bernhard Schölkopf \\
  ETH Zürich \\
  \texttt{bs@tuebingen.mpg.de} \\
  \\\And
  Andreas Geiger \\
  University of Tübingen \\
  \texttt{a.geiger@uni-tuebingen.de} \\
}

\begin{document}

\compacttitlespacing

\setlength{\abovedisplayskip}{4pt plus 0.5pt minus 0.5pt}
\setlength{\belowdisplayskip}{4pt plus 0.5pt minus 0.5pt}
\setlength{\abovedisplayshortskip}{0.5pt plus 0.5pt}
\setlength{\belowdisplayshortskip}{0.5pt plus 0.5pt}

\maketitle
\begin{abstract}
ARC tests in-context rule induction: given a few input-output demonstrations, a model must infer the hidden rule and apply it to a new query. While many approaches express ARC rules through language, code, or symbolic programs, ARC itself is visual-symbolic: rules appear as grid transitions over objects, colors, shapes, and spatial relations. We introduce Loop-OWM, an object-centric world-modeling architecture that learns these rules as composable transitions over structured states. It combines color-prototype slots, demonstration-conditioned task summaries, and a looped transition model with dense propagation and slot-conditioned correction. On both ARC-1 and ARC-2, Loop-OWM outperforms non-looped and looped baselines with comparable or fewer parameters. These results suggest that ARC rules can be learned not only as language descriptions or searched programs, but also as transitions over visual-symbolic world states.
\end{abstract}

\section{Introduction}

The Abstraction and Reasoning Corpus (ARC) was introduced as a benchmark for studying abstract reasoning under limited supervision: each task provides only a few input-output demonstrations, and the model must infer the hidden rule and apply it to a new query input. This setting has naturally attracted language-centric and program-centric approaches, where the demonstrations are interpreted as prompts for rule induction, search, or program synthesis. However, the task interface of ARC is not linguistic. The rules are expressed through transitions of discrete 2D grids: objects are moved, copied, recolored, completed, counted, or related through spatial configurations. ARC therefore sits at an interesting boundary between language-like reasoning and visual-symbolic state transition.

Recent visual approaches show that this visual-symbolic interface can be modeled directly by representing ARC examples as grid images or visual canvases. However, they still largely treat ARC as direct grid-to-grid prediction: the model is asked to produce the target grid, while the intermediate structure of the rule -- what objects are present, how they change, and how the observed transition should apply to a new state -- remains implicit.
This is limiting because many ARC tasks are naturally object-indexed. We therefore ask: can ARC rules be learned as composable object-aware transitions over structured grid states, and rolled out through iterative updates?

In this paper, we study ARC from the perspective of \textit{object-centric world modeling}. In this view, each grid is a visual-symbolic world state, each demonstration pair provides evidence of a state transition, and each task specifies a state transition rule that should generalize from given pairs to news query examples. 
Solving an ARC task then requires more than predicting an output grid:  
the model must interpret the current state, identify reusable object-level structure, infer how the state changes from demonstrations, and execute the inferred transition on a new input. 
This formulation connects ARC to in-context rule induction: the demonstration pairs act as a prompt, but the induced rule is grounded in structured visual states rather than expressed purely in natural language or code.

We propose \textbf{Loop-OWM}, a composable architecture built around this formulation. The model consists of three conceptual components. First, an \textit{object-centric grid interpreter} converts ARC grids into structured visual-symbolic states, exposing candidate entities through slots. Second, a \textit{task encoder} reads the demonstration inputs, outputs, and transition-related features to infer a task-specific rule representation. Third, a \textit{looped transition model} applies the inferred rule to the query state through iterative updates. The loop is not used merely as additional computation depth; rather, it implements a rollout mechanism in which the model repeatedly updates an internal state under a task-conditioned transition rule.

This design allows us to separate high-level modeling principles from lower-level architectural choices. Loop-OWM is not a single monolithic predictor, but a structured design space for studying how neural models can learn visual-symbolic transition rules. In our final instantiation, each recurrent update is implemented through two complementary branches: one branch models \textit{dense grid-level} state propagation, while the other injects \textit{object-aware corrections} through slot-conditioned representations. 
We evaluate design choices through controlled comparisons with non-looped and looped ViT baselines, together with ablation studies on object-aware updates, task-summary tokens, transition supervision, and recurrent rollout. 
Beyond improving ARC prediction, we hope this study contributes to a broader view of abstract reasoning: ARC rules need not be represented only as language descriptions or searched programs, but can also be learned as composable transitions over structured world states.

\section{Related Work}

\subsection{ARC as Rule Induction}
ARC was introduced to measure skill-acquisition efficiency and abstract reasoning from
limited experience \citep{chollet2019measure}. Each task provides only a few
input-output demonstrations, requiring the solver to infer a hidden transformation rule
and apply it to a new query. This has motivated symbolic, program-centric, and more
recently LLM-based approaches that describe, search, or synthesize candidate programs
for ARC tasks \citep{wind2020dsl, ouellette2024efficient, singhal2024conceptsearch}. Such methods exploit the discrete and
compositional structure of ARC, but often rely on explicit program spaces, hand-designed
primitives, execution-guided search, or language-mediated reasoning. Recent analyses
also suggest that LLMs can identify high-level concepts but still struggle with
compositional execution of spatial transformations \citep{lee2024reasoning}. 
We study a complementary direction: learning ARC rules as latent transitions over
visual-symbolic grid states, rather than expressing them explicitly as language or code.

\subsection{Neural Visual Models for ARC}
Recent work shows that ARC's visual interface can be modeled directly. Vision-centric
ARC methods represent tasks as spatial canvases and train neural architectures for
grid-to-grid translation \citep{hu2025varc}. Looped visual transformers further introduce
recurrent computation, repeatedly applying shared blocks to increase effective reasoning
depth without increasing parameter count \citep{shu2026loopvit, saunshi2025reasoning, geiping2025scaling, schwarzschild2021learnalgorithmgeneralizingeasy}. These approaches
demonstrate that visual modeling and iterative computation are useful inductive biases
for ARC. However, they primarily treat the task as output-grid prediction. In contrast,
Loop-OWM formulates ARC as demonstration-conditioned state transition learning: the
model must infer what objects are present, how they change, and how the observed
transformation should be applied compositionally to a new state.

\subsection{Object-Centric and World Models}
Many ARC rules are naturally object-indexed: transformations often operate on shapes, colors, connected components, and spatial relations rather than independent grid cells
\citep{chollet2019measure,xu2024llmsabstractionreasoningcorpus}. This makes object-centric representation learning relevant to ARC. Slot-based methods provide a mechanism for exposing an entity-like structure from dense visual features \citep{burgess2019monet, locatello2020slot,kirilenko2024objectcentric}, while object-centric video and world models show that such representations can support dynamics and relational reasoning \citep{zadaianchuk2023videosaur, wu2023slotformer,
nam2026causaljepa}. Our setting differs from video or control domains: ARC provides no natural temporal trajectories, so the transition rule must be inferred from a few input-output demonstrations. 
Loop-OWM therefore treats the demonstration pairs themselves as evidence for a
task-specific dynamics model.

\section{Problem Formulation}

An ARC task consists of a small set of input-output demonstrations
\begin{equation}
\mathcal{D} = \{(x^{(i)}, y^{(i)})\}_{i=1}^{N},
\end{equation}
and a query input \(x \in \{0,\ldots,C-1\}^{H \times W}\) where each grid is a discrete visual-symbolic state, \(C\) counts the valid ARC colors. The goal is to predict the corresponding output grid \(y\) by inferring the task-specific rule expressed by the demonstrations. 
A direct prediction model treats this problem as
\begin{equation}
\hat{y} = f_\theta(x, \mathcal{D}),
\end{equation}
where the demonstrations act as conditioning information for grid-to-grid prediction.

In contrast, we formulate ARC as demonstration-conditioned state transition learning. Each demonstration pair \((x^{(i)}, y^{(i)})\) provides evidence of how a state changes under the hidden task rule. The model must infer a latent transition rule \(z_{\mathcal{D}}\) from the demonstrations and apply it to the query state:
\begin{equation}
z_{\mathcal{D}} = \psi_\theta^{\mathrm{task}}(\mathcal{D}), 
\quad
\hat{y} = \mathrm{Rollout}(x, z_{\mathcal{D}}).
\end{equation}
This separates rule induction from rule execution: the task encoder \(\psi_{\theta}^{\mathrm{task}}(\cdot)\) infers what transition should be applied, while the rollout model applies the inferred rule to a new state.
We use the term \textbf{\textit{world model}} in this \textbf{\textit{restricted sense}}: a model of structured ARC states and the task-conditioned transitions between them, rather than a model of physical dynamics or embodied control.

\noindent \textbf{Object-aware state abstraction. }
We introduce an object-aware state abstraction. 
Given a grid \(x\), we first map each cell color to an embedding
using a shared color embed table \(E\in\mathbb{R}^{C \times d}\). The resulting feature map is patchified with a \(\mathrm{p} \times \mathrm{p}\) convolution~\citep{dosovitskiy2021an} and augmented with \(2D\) positional embeddings, yielding dense grid-state tokens
\begin{equation}
H = \phi^{\mathrm{grid}}(x) \in \mathbb{R}^{L \times d},
\quad
L = (H/\mathrm{p})(W/\mathrm{p}).
\end{equation}
These dense tokens preserve fine-grained spatial information and serve as the grid-level
state representation used by the transition model.

From the dense grid-state tokens, we extract a set of $K$ object slots
\begin{equation}
S = \{s_k\}_{k=1}^{K}, \quad s_k \in \mathbb{R}^{d},
\end{equation}
using a \textit{Slot Attention}~\cite{locatello2020slot} module \(\mathrm{S}_\theta(\cdot)\): the slots attend competitively to the dense tokens and are iteratively updated through attention, recurrent refinement, and a residual MLP. 

The slots are not required to be explicit symbolic objects or perfect segmentation masks.
Instead, they serve as a \textit{structured interface} through which the model can expose
candidate entity-level information to the task encoder and transition model.

\noindent \textbf{Looped rule execution.} 
Given the inferred task rule \(z_{\mathcal{D}}\), the query grid is initialized as a latent state
\begin{equation}
H^{0} = \phi^{\mathrm{grid}}(x).
\end{equation}
The model then applies a shared Transformer-based~\citep{vaswani2023attentionneed} transition module for \(L\) recurrent steps:
\begin{equation}
H^{t+1} = F_\theta(H^t, z_{\mathcal{D}}),
\quad t = 0,\ldots,T-1.
\end{equation}
The final state is decoded into logits $\hat{Y}$ over the ARC color vocabulary, 
\begin{equation}
    \hat{Y} = \mathrm{Dec}_\theta(H^{T}), 
\end{equation}
which defines a categorical distribution over colors.
\begin{equation}
    p_\theta(c \mid u,v) = \mathrm{softmax}\left[ \hat{Y}_{(u,v)}\right]_c.
\end{equation}
Given the ground-truth output grid $y$, we define a standard cross-entropy loss over valid cells:
\begin{equation}
\label{loss:grid}
\mathcal{L}_{\mathrm{grid}} 
= -\frac{1}{|\mathcal{V}|} \sum_{(u,v)\in \mathcal{V}} \log p_\theta(y_{(u,v)} \mid u,v),
\end{equation}
where \(\mathcal{V}\) denotes the set of valid non-padding cells.

The loop should not be understood merely as extra depth. It provides a rollout mechanism: the same inferred rule can be repeatedly applied to an evolving internal state, allowing ARC prediction to be framed as iterative state transition rather than one-shot image translation.
For simplicity, we use \(\theta\) to collectively denote all learnable parameters.

\begin{figure}[t]
\centering
\includegraphics[width=\linewidth]{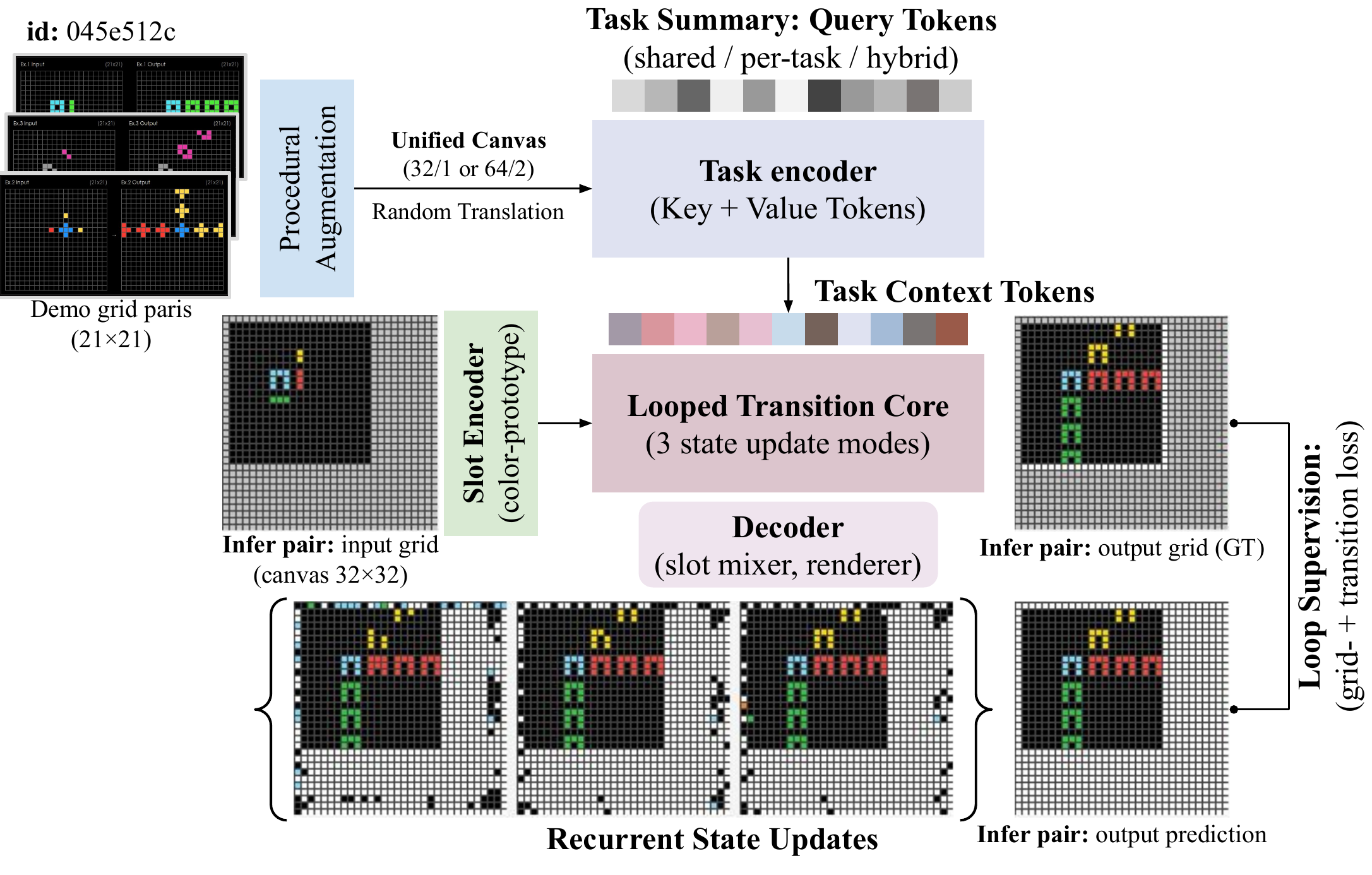}
\vspace{-5mm}
\caption{
\textbf{Loop-OWM overview.}
Demonstration pairs are augmented and encoded into task context tokens, which condition an object-aware transition model to recurrently update the query state and decode the final prediction under grid and transition supervision.
}
\vspace{-5mm}
\label{fig:overview}
\end{figure}

\section{Method} 
Loop-OWM instantiates the formulation above with four learnable modules: an object-centric interpreter \(\mathrm{S}_\theta(\cdot)\), a task encoder \(\psi_{\theta}^{\mathrm{task}}(\cdot)\), a looped transition model \(F_\theta(\cdot)\) with decoding module \(\mathrm{Dec}_\theta(\cdot)\). 
An overview of Loop-OWM is shown in Figure~\ref{fig:overview}.

In this section, we focus on the \textbf{\textit{design choices}} that instantiate the formulation: how grids are interpreted as object-aware states, how demonstrations are summarized into task representations, how the looped transition module updates the
query state, and how different transition branches and supervision signals are combined
in the final model.

\subsection{Color-Prototype Slot Initialization}
A key design choice is how slots are initialized. Instead of sampling free Gaussian slots per image, we initialize each slot from a \textit{color prototype}. Specifically, the initial seed of slot \(i\) is given by
\begin{equation}
s_i^{0} = E_i + b_i \in \mathbb{R}^d
\end{equation}
where \(E_i\) is a row of the embedding table $E$ 
and \(b_i\) is a learnable slot-specific bias, \(b_i = \mu + \sigma \cdot \epsilon_i\) 
with $\mu, \sigma \in \mathbb{R}^d$ shared learnable bias and log-variance, and $\epsilon_i \sim \mathcal{N}(0,I)$ resampled at every forward pass. 
Special symbols such as padding or ignored tokens are not used as slot prototypes. When $K = C$, slot $i$ is seeded by $E_i$ directly; if $K$ is a positive integer multiple of $C$, we tile the table and each color anchors $K/C$ slots that share the same seed.

This color-prototype initialization binds slot identity to the shared ARC color palette.
As a result, the same slot index has a consistent color anchor across demonstration pairs and query grids. This is useful for ARC because color is not merely a visual attribute; it often functions as a symbolic identifier for objects, regions, or roles within a task. The prototype slots therefore provide a stable object-aware interface while still allowing the Slot Attention updates to adapt each slot to the spatial context of a particular grid.

\noindent \textbf{Stationary Color Anchors.}
Since the same color embedding table \(E\in\mathbb{R}^{C \times d}\) is used both to encode grid cells and to initialize slot prototypes, unconstrained updates to \(E\) would move the color anchors
during training. This can blur the intended correspondence between slot identity and
ARC color identity. To keep the prototype anchors stable, we initialize \(E\) as an
\textit{orthogonal} row frame and \textit{freeze} it throughout training:
\begin{equation}
E E^{\top} \approx d \cdot \mathbf{I}_C ,
\end{equation}
up to the subset of rows used for valid ARC colors. The scaling by \(d\) corresponds
to using a gain of \(\sqrt{d}\), so that each color embedding has norm \(\sqrt{d}\).
This keeps the input magnitude comparable to a standard embedding layer and avoids
changing the scale seen by downstream patch and positional layers. 
Empirically, we find that freezing \(E\) stabilizes the training of the
looped transition model \(F_\theta(\cdot)\).

\subsection{Complementary Transition Module}
\label{subsec:update_modes}
The module \(F_\theta(\cdot)\) serves as the transition operator of the ARC world model. Given the current state \(H^t \in \mathbb{R}^{L \times d}\) and the inferred task context \(z_{\mathcal{D}}\), it predicts both how information should be transported across grid positions and what slot-conditioned content should be written back into the state.

The update is decomposed into two complementary branches. The first branch predicts a dense \textit{patch-to-patch transition} operator
\begin{equation}
P^t \in \mathbb{R}^{L \times L},
\end{equation}
which routes information over grid positions. The second branch writes \textit{slot-conditioned} content back to the patch level, producing an update
\begin{equation}
R^t \in \mathbb{R}^{L \times d}.
\end{equation}
Together, these branches allow the model to represent both spatial propagation and object-aware content modification. In practice, they are implemented with a shared task-conditioned cross-attention backbone and separate prediction heads..

\noindent \textbf{Slot-Conditioned Update.}
To map object-level information back to the grid state, inspired by VideoSAUR~\citep{zadaianchuk2023videosaur}, we introduce a slot-to-grid mixer \(\mathcal{M}_\theta^{\mathrm{s2g}}(\cdot)\) as part of the decoding module \(\mathrm{Dec}_\theta(\cdot)\). 
At each loop step, 
the mixer maintains a set of learnable positional patch queries \(\mathbf{Q} = \{q_j\}_{j=1}^{L}\), where \(q_j \in \mathbb{R}^{d}\) corresponds to one grid patch and is $t$-independent. 
Each patch query \(q_j\) attends to the current slots $S^t=\{s_i^t\}^K_{i=1}$ and reconstructs a slot-conditioned patch feature:
\begin{equation}
\tilde{h}_j^t
=
\sum_{i=1}^{K}
\mathrm{softmax}_{i}
\left(
\frac{
\langle W_\mathbf{Q} q_j, W_\mathbf{K} s_i^t \rangle
}{
\sqrt{d}
}
\right) 
W_\mathbf{V} s_i^t .
\end{equation}
We denote the reconstructed patch features as
\begin{equation}
\tilde{\mathbf{h}}^{t} 
:= \{\tilde{h}_j^t\}_{j=1}^{L}
:= \mathcal{M}_\theta^{\mathrm{s2g}}(\mathbf{Q}, S^t)
\in \mathbb{R}^{L \times d} 
\end{equation} 
and pass them through a lightweight MLP renderer to produce the slot-conditioned update 
\begin{equation}
R^t := \mathrm{MLP}_\theta(\tilde{\mathbf{h}}^{t}).
\end{equation}
This branch serves as the channel through which object-level information can write new or corrected content into the grid state.

\noindent \textbf{State update modes.} 
We study three ways of composing \(P^t\) and \(R^t\) into the next state: 

\noindent \textbf{(1) Slot-only update.}
The next state is fully generated by the slot-to-grid mixer:
\begin{equation}
H^{t+1} = R^t .
\end{equation}
This mode is closest to the 
formulation used in object-centric video models~\citep{zadaianchuk2023videosaur, nam2026causaljepa}, where \(P^t\) is only trained through auxiliary transition supervision and does not affect the forward prediction.

\noindent\textbf{(2) Transport-only update.}
The next state is produced only by dense transport:
\begin{equation}
H^{t+1} = P^t H^t .
\end{equation}
This forces the model to express the transition purely as routing over existing patch features. 
While suitable for permutation-like transformations, this mode is too restrictive for ARC rules that require writing new content, such as recoloring, filling, or completing shapes.

\noindent\textbf{(3) Complementary update.}
The last tests their complementarity:
\begin{equation}
H^{t+1} = P^t H^t + \alpha^t \cdot R^t .
\end{equation}
where \(\alpha^t\) is a learnable gate parameter. 
The two paths therefore play complementary roles:
\(P^t\) keeps the dense spatial propagation in the forward computation, while
\(R^t\) provides the object-aware update to express content changes that cannot be captured by transport alone. 
We ablate different updating modes in Section~\ref{subsec:ablatiob}.

\subsection{Task Encoder with Summary Tokens}
\label{subsec:task_summary}

We use the task encoder $\psi_\theta(\cdot)$ to summarize the demonstration set $\mathcal{D}$ into a small set of context tokens that condition the looped transition model $F_\theta(\cdot)$ at every step 
\footnote{In implementation, at each loop step, a loop-index embedding is concatenated with \(z_{\mathcal{D}}\) along the sequence dimension before being used to condition the transition model
\(F_\theta(\cdot)\).}.

Given the encoded demo-pair features, we construct a sequence of \textit{demo-pair tokens} that represent the input/output states and transition-related features for each example. 
Depending on the update mode (introduced in Section~\ref{subsec:update_modes}), these pair tokens may be defined over slot-level representations or patch-level grid features. 
A learnable \textit{task-summary query bank} then cross-attends to these demo pair tokens, producing the task embedding $z_{\mathcal{D}}$. More implementation details for the task encoder are provided in \ref{app:task_encoder_impl}.

The main design choice is how to parameterize the task-summary queries. Consider three variants:

\noindent \textbf{(1) Shared-only Summary.}
The summary queries are shared across all tasks. Conditioning on a particular task therefore comes entirely from cross-attention to that task’s demonstration tokens. This encourages amortized rule induction: the same query bank must learn a general procedure for reading demonstrations and extracting a task rule.

\noindent \textbf{(2) Per-task Summary.}
The summary queries can instead be retrieved from a task-specific lookup table. For a training task $m$, the encoder uses a learned table entry $E^\mathrm{task}_m$ as the initial summary queries. This gives the model task-specific memory and may help when the same training tasks are revisited many times. However, it also ties part of the representation to the training task-id space, so unseen tasks require either newly initialized task entries and test-time adaptation.

\noindent \textbf{(3) Hybrid Summary.}
The hybrid variant combines shared global summary tokens with task-specific summary tokens. This design separates two roles: shared tokens provide a task-agnostic interface for reading demonstrations, while task-specific tokens provide additional capacity for storing task-dependent biases. During training, task-token dropout can be applied so that the model cannot rely exclusively on task identifiers and must still use the demonstration readout.

These variants instantiate different points in the trade-off between amortized rule induction and task-specific memory. The shared variant is most compatible with unseen task generalization and cross-benchmark transfer; the per-task variant provides stronger task-specific capacity; and the hybrid variant attempts to combine both. We therefore treat the choice of task-summary parameterization as an empirical design question and compare the variants in the ablation study in Section~\ref{subsec:ablatiob}.

\begin{table*}[t]
\centering
\vspace{-5mm}

\begin{minipage}[t]{0.49\textwidth}
\vspace{0pt}
\centering
\small
\setlength{\tabcolsep}{0.5pt}
\renewcommand{\arraystretch}{1.12}
\begin{tabularx}{\linewidth}{@{}>{\raggedright\arraybackslash}X
>{\centering\arraybackslash}p{0.15\linewidth}
>{\centering\arraybackslash}p{0.14\linewidth}
>{\centering\arraybackslash}p{0.14\linewidth}@{}}
\toprule
System & \#Params & \textbf{ARC-1} & \textbf{ARC-2} \\

\specialrule{\lightrulewidth}{0pt}{0pt}
\rowcolor{gray!15}
\multicolumn{4}{@{}l@{}}{\textit{ViT}} \\

VARC~\citep{hu2025varc} 
& 18M 
& 54.5 
& 8.3\\

VARC (ensemble) 
& 73M 
& 60.4 
& 11.1 \\

\specialrule{\lightrulewidth}{0pt}{0pt}
\rowcolor{gray!15}
\multicolumn{4}{@{}l@{}}{\textit{recurrent models}} \\

HRM~\citep{hrm} 
& 27M 
& 40.3 
& 5.0 \\

TRM~\citep{trm} 
& 7M 
& 44.6 
& 7.8 \\

\specialrule{\lightrulewidth}{0pt}{0pt}
\rowcolor{gray!15}
\multicolumn{4}{@{}l@{}}{\textit{recurrent ViT}} \\

LoopViT-S~\citep{shu2026loopvit} 
& 3.8M 
& 60.1 
& 10.0 \\

LoopViT-M 
& 11.2M 
& 63.8 
& 11.5 \\

LoopViT-L 
& 18M 
& 65.8 
& 14.2 \\

\specialrule{\lightrulewidth}{0pt}{0pt}
\textbf{Loop-OWM 32/1 (Ours)}
& 10.42M 
& \textbf{67.3} 
& \textbf{20.2} \\

\textbf{Loop-OWM 64/2 (Ours)}
& 10.63M 
& \textbf{68.5} 
& \textbf{22.5} \\

\bottomrule
\end{tabularx}
\end{minipage}
\hfill
\begin{minipage}[t]{0.49\textwidth}
\vspace{0pt}
\centering
\footnotesize
\setlength{\tabcolsep}{0.2pt}
\renewcommand{\arraystretch}{1.21}
\begin{tabularx}{\linewidth}{@{}
>{\raggedright\arraybackslash}X
>{\centering\arraybackslash}p{0.18\linewidth}
>{\centering\arraybackslash}p{0.21\linewidth}
>{\centering\arraybackslash}p{0.13\linewidth}
>{\centering\arraybackslash}p{0.13\linewidth}@{}}
\toprule
System 
& Type 
& \shortstack{Cost/task(\$)}
& ARC-1 
& ARC-2 \\

\specialrule{\lightrulewidth}{0pt}{0pt}
\rowcolor{gray!15}
\multicolumn{5}{@{}l@{}}{\textit{large language models (LLMs)}} \\

\textcolor{gray}{Deepseek V3.2}
& \textcolor{gray}{Base LLM}
& \textcolor{gray}{0.080}
& \textcolor{gray}{57.0} 
& \textcolor{gray}{4.0} \\

\textcolor{gray}{o3-Pro (High)} 
& \textcolor{gray}{CoT+Syn.}
& \textcolor{gray}{7.55} 
& \textcolor{gray}{59.3} 
& \textcolor{gray}{4.9} \\

\textcolor{gray}{o3 (Preview, Low)} 
& \textcolor{gray}{CoT+Syn.}
& \textcolor{gray}{200.00} 
& \textcolor{gray}{75.7} 
& \textcolor{gray}{4.0} \\

\textcolor{gray}{Gemini 3.1 Pro (Pre.)} 
& \textcolor{gray}{CoT}
& \textcolor{gray}{0.962} 
& \textcolor{gray}{98.0} 
& \textcolor{gray}{77.1} \\

\textcolor{gray}{GPT-5.5 (xHigh)} 
& \textcolor{gray}{CoT}
& \textcolor{gray}{1.87} 
& \textcolor{gray}{95.0} 
& \textcolor{gray}{85.0} \\

\specialrule{\lightrulewidth}{0pt}{0pt}
\rowcolor{gray!15}
\multicolumn{5}{@{}l@{}}{\textit{human results}} \\

\textcolor{gray}{Human Panel} 
& \textcolor{gray}{-} 
& \textcolor{gray}{-}
& \textcolor{gray}{98.0} 
& \textcolor{gray}{100.0} \\

\textcolor{gray}{Avg. Human} 
& \textcolor{gray}{-} 
& \textcolor{gray}{-}
& \textcolor{gray}{60.2} 
& \textcolor{gray}{-} \\

\bottomrule
\end{tabularx}
\raggedright
\tiny
\textit{Note.} System type describes the inference strategy: \\
\textit{Base LLM} represents single-shot inference from standard language models; \\
\textit{CoTs} represents chain-of-thought reasoning; \\
\textit{CoT+Syn.} denotes CoT reasoning with additional synthesized programs or candidate solutions.
\end{minipage}

\vspace{-2mm}
\caption{
\textbf{System-level Comparison} on ARC-1 and ARC-2.
\textbf{(Left)} Compact neural ARC systems. 
Loop-OWM uses two-stage training: offline procedural augmentation followed by task-specific test-time training with task-variant augmentation. 
ARC-2 results use ARC-1-to-2 transfer initialization from the checkpoint trained on ARC-1 for 200 offline epochs. 
We report \textit{pass@2 mean accuracy} over 5 runs, using 30 random test-time views per task. Loop-OWM performs 4 recurrent steps; parameter counts exclude per-task summary tables.
\textbf{(Right)} LLM-based systems and human references~\cite{arc-agi,arc-human}. LLM results are from the
ARC-AGI leaderboard and use different inference strategies, including base LLM
inference, chain-of-thought reasoning, and synthesized candidate solutions.
}
\vspace{-3mm}
\label{tab:arc_comparison}
\end{table*}

\subsection{Composed Transition across Loops}
\label{sec:composed_transition}

How the transition matrix \(P^t\) is computed and supervised across loops is an orthogonal design choice. 
We separate \textit{two axes}: (i) what target the predicted transition
is compared against, and (ii) which transition quantity is supervised. For the first
axis, we compare raw patch-feature similarity with a slot-mediated similarity target.
For the second axis, we compare supervising an individual transition matrix with
supervising the composed transition induced by the full looped rollout.

\noindent \textbf{(1) Dense vs. Slot-mediated Transition.}
Following VideoSAUR~\citep{zadaianchuk2023videosaur},
a natural target for dense patch-wise transport is the softmax-normalized cosine similarity
between destination ($\cdot^{\mathrm{dst}}$) and source ($\cdot^{\mathrm{src}}$) patches $q$ and $p$ :
\begin{equation}
\mathcal{P}^{\mathrm{raw}}_{q,p}
=
\mathrm{softmax}_{p}
\left(
\frac{
\langle \hat{f}^{\mathrm{dst}}_{q}, \hat{f}^{\mathrm{src}}_{p} \rangle
}{
\tau
}
\right),
\end{equation}
where \(\hat{f}\) denotes a \(\ell_2\)-normalized patch feature and \(\tau\) is a
temperature. 

Empirically, we found this target can be \textit{biased} towards \textit{identity transport}: patch features include position-dependent information, source and destination patches at the same coordinate can become similar simply because they share
the same position, even when the task rule requires nontrivial content movement.

To reduce this positional identity bias, we use a \textit{slot-mediated} transition target.
We pass both the source and destination grids through the same slot encoder \(\mathrm{S}_\theta(\cdot)\) with shared color-prototype slot seeds, obtaining slot-membership masks
\begin{equation}
M^{\mathrm{src}}, M^{\mathrm{dst}} \in \mathbb{R}^{L \times K},
\end{equation}
where \(M^{\mathrm{*}}_{j,k}\) is the membership of a src/dst patch \(j\) to slot
\(k\). We compute the slot-mediated similarity by marginalizing over the shared
slot index: 
\begin{equation}
\mathrm{sim}^{\mathrm{slot}}_{q,p}
=
\sum_{k=1}^{K}
M^{\mathrm{dst}}_{k,q}
M^{\mathrm{src}}_{k,p},
\end{equation}
and define the transition target by normalizing over source patches:
\begin{equation}
\mathcal{P}^{\mathrm{slot}}_{q,p}
=
\mathrm{softmax}_{p}
\left(
\frac{
\mathrm{sim}^{\mathrm{slot}}_{q,p}
}{
\tau
}
\right).
\end{equation}
This target asks whether a destination patch and a source patch are assigned to the
same color-prototype slot, rather than whether their raw patch embeddings are similar.
It therefore reduces positional identity bias and provides a transport target aligned
with the object-centric abstraction used by the model. 

Note that the color-prototype initialization provides a \textit{shared slot coordinate system} across source and destination grids: unlike randomly initialized slots, whose ordering is permutation-symmetric, prototype slots are anchored to the ARC color palette and are therefore more consistently aligned across grids. 
In practice, we use this \textit{slot-mediated target} in the main model.

\noindent \textbf{(2) Final-loop vs. Composed Transition.}
The second design choice concerns which transition quantity should be supervised. A
straightforward option is to attach the transition loss only to the final-step transition
matrix 
\begin{equation}\hat{P}^\mathrm{final} = \hat{P}^{T-1}.\end{equation} 
This supervises the last routing operation before decoding, but does not directly constrain the effective transport induced by the full looped rollout.

When applying the transition module recurrently, the overall transport from the initial query state to the final state is given by the composition of all per-step transition matrices.  
With the update convention \(H^{t+1} = P^t H^t ~(+ R^t)\), the effective transition after $T$ steps is
\begin{equation}
\hat{P}^{\mathrm{total}}
=
\hat{P}^{T-1}
\hat{P}^{T-2}
\cdots
\hat{P}^{0}.
\end{equation}
We supervise \(\hat{P}^{*}\) using a transition target computed between the query input grid and the ground-truth output grid:
\begin{equation}
    \mathcal{L}_{\mathrm{trans}}
    =
    -
    \frac{1}{L}
    \sum_{q=1}^{L}
    \sum_{p=1}^{L}
    \mathcal{P}^*_{q,p}
    \log
    \left(
    \hat{P}^*_{q,p} + \epsilon
    \right),
\end{equation}
where \(\mathcal{P}^*_{q,p}\) is the transition target chosen (either raw or slot), and \(\hat{P}^*_{q,p}\) is the supervised prediction. 

Supervising \(\hat{P}^{\mathrm{total}}\) aligns the transport objective with the actual multi-step routing experienced by the query state. In contrast to supervising only $\hat{P}^\mathrm{final}$, the composed objective discourages any single step from collapsing to a trivial identity solution unless the remaining steps compensate for the full input-to-output transport. 
In practice, we supervise the \textit{composed total transition} in our main setting. 

\noindent \textbf{Overall objective.} The full training loss is:
\begin{equation}
    \mathcal{L} = \mathcal{L}_\text{grid} + \lambda \, \mathcal{L}_\text{trans},
    \label{eq:total_loss}
\end{equation}
where $\lambda$ controls the transition loss weight.

\section{Experiments}


\noindent \textbf{Datasets.} We evaluate Loop-OWM on ARC-1 and ARC-2~\citep{chollet2019measure, arc-agi}. Since the original ARC files provide only a few demonstrations per task, training directly on the raw files gives very sparse supervision for learning neural transition models. We therefore use procedural augmentation to generate additional input-output pairs for the training tasks, while keeping the official evaluation tasks out. More implementation and configurations details are provided in Appendix~\ref{app:param_count}. 

\subsection{Two-stage Training}
We adopt a two-stage training pipeline consisting of offline pretraining and test-time
adaptation.  
This protocol is natural for ARC because each test task may instantiate a
rule that is not directly covered by the offline training distribution. The offline
stage trains general visual-symbolic machinery for grid interpretation, object-centric
abstraction, and transition modeling, while the test-time stage specializes this
machinery to the rule specified by the demonstrations of the current task. This separates
general mechanism learning from task-specific rule specialization, matching the few-shot
nature of ARC.

\noindent \textbf{Offline procedural augmentation. }
For ARC-1, we augment the 400 training tasks with RE-ARC \citep{hodel2024rearc}, a procedural reimplementation of the ARC-1 training tasks. 
For each original training task, RE-ARC can generate fresh input-output pairs by sampling from the underlying task program. Following \citep{hu2025varc}, we generate approximately 1000 pairs per task, allowing the model to train with dense task-level supervision rather than only the few examples provided in the raw ARC files. This augmentation is applied only to the training split; the evaluation split is held out and never used for generation or training. 

For ARC-2, we use ARC-GEN \citep{arcgen2025}, a mimetic procedural generator for ARC. Among the 1000 ARC-2 training tasks, 891 task ids are covered by ARC-GEN: 500 correspond to ARC-2-specific task instances, while the
remaining 391 are overlapping with ARC-1.
We generate up to 1000 pairs per covered task. Tasks whose rejection sampler exceeds a per-call time budget are dropped; this affects 16 tasks, all from the
ARC-2-only subset. None of the 120 ARC-2 evaluation tasks have a corresponding generator, so the offline augmentation is leakage-free.

\noindent \textbf{Test-time augmentation.}
We follow the the \textit{task-variant augmentation} strategy introduced in VARC~\citep{hu2025varc}:
for each original test task, we create 50 augmented task variants via rotations, flips, or color permutations. Resuming the offline-trained checkpoint, we further train the model on the demonstrations of the current test task and their augmented variants. 
For inference, we evaluate the adapted model on \textit{multi-augmented views}, invert the transformations on the predicted grids, and consolidate the resulting candidates by \textit{majority voting}. 
We use \(30\) random views per task and report the mean accuracy over \(5\) independent runs.

\noindent \textbf{ARC-1 to ARC-2 transfer. }
For ARC-2, we additionally study cross-benchmark initialization from ARC-1.
Specifically, we first train Loop-OWM on ARC-1 for 200 offline epochs, and then use this checkpoint to initialize offline training on ARC-2. 
Only task-independent parameters are transferred; task-id-dependent parameters, such as per-task summary tables, are re-initialized for ARC-2. 
This protocol tests whether the model can transfer general visual-symbolic machinery across ARC benchmarks. Detailed training curves and analysis are provided in
Appendix~\ref{app:cross_benchmark_transfer}.

\begin{figure}[t]
\centering
\begin{subfigure}{\linewidth}
\centering
\includegraphics[width=\linewidth]{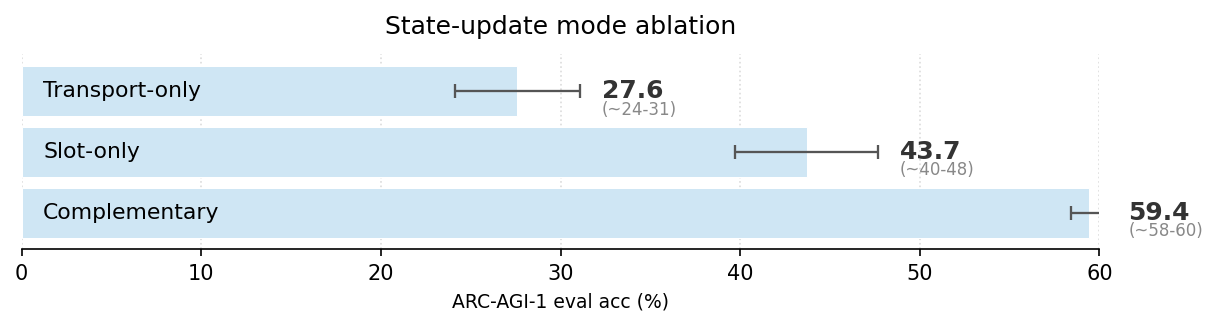}
\caption{\footnotesize{State-update composition.}}
\label{fig:update_ablation}
\end{subfigure}

\vspace{0.5em}

\begin{subfigure}{\linewidth}
\centering
\includegraphics[width=\linewidth]{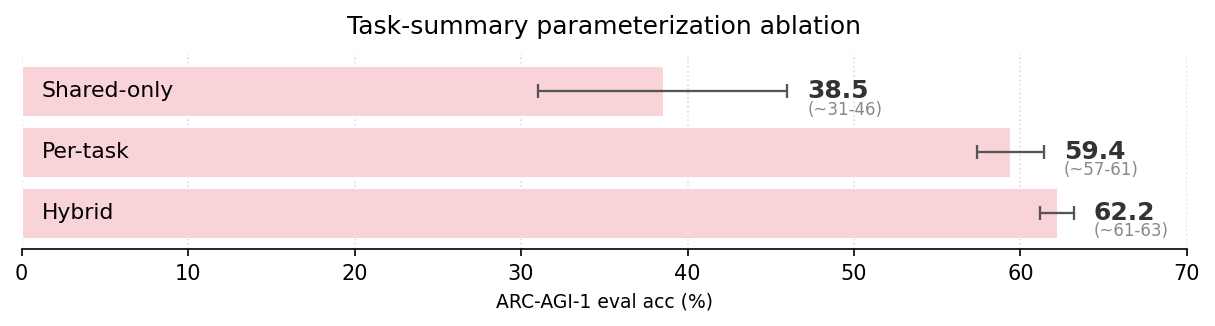}
\caption{\footnotesize{Task-summary parameterization.}}
\label{fig:summary_ablation}
\end{subfigure}

\vspace{-2mm}
\caption{
Ablation results on the ARC-1 evaluation set using the \(32/1\) configuration.
All variants use \(100\) offline training epochs and \(100\) test-time training epochs,
and are evaluated with a reduced multi-view budget of \(10\) random views over \(2\)
independent runs. 
}
\vspace{-5mm}
\label{fig:ablation}
\end{figure}

\subsection{System-Level Comparison}

We compare Loop-OWM with prior neural systems, LLMs, recurrent methods, and human-level references on ARC-1 and ARC-2. 
As shown in Table~\ref{tab:arc_comparison} \textbf{(Left)}, Loop-OWM achieves the strongest results among compact neural systems. 
Compared with recurrent baselines~\citep{hrm, trm, shu2026loopvit}, Loop-OWM achieves substantially higher performance while remaining in a compact parameter regime. 

Table~\ref{tab:arc_comparison} \textbf{(Right)} also highlights a different trade-off from LLM-based systems. 
Recent LLMs can obtain strong performance on ARC, especially when using chain-of-thought (CoT) reasoning with optional synthesized candidate solutions. However, this performance often comes with substantially higher inference cost. 
In contrast, Loop-OWM is a compact ViT-based model trained to operate directly on grid-structured inputs, without relying on external language reasoning, program synthesis, or LLM-scale candidate search. 
This suggests that explicitly modeling object-centric recurrent transitions can provide an effective inductive bias for ARC-style visual reasoning.

Finally, the gap between ARC-1 and ARC-2 remains significant for all neural systems, indicating that ARC-2 continues to pose a substantially harder generalization challenge. 
Nevertheless, Loop-OWM narrows this gap relative to prior compact neural models, suggesting that recurrent object-world modeling is a promising direction for scaling vision-based ARC solvers.

\subsection{Ablation Study}
\label{subsec:ablatiob}

We ablate the two design choices most directly tied to our modeling hypothesis:
how the looped transition module composes dense and object-aware updates,
and how the task encoder parameterizes the summary tokens used for rule induction.
All variants are conducted on ARC-1 using a \(32 \times 32\) canvas with patch size p \(=1\)) and the same two-stage training protocol: \(100\)-ep offline training followed by \(100\)-ep test-time training.
We use \(10\) random views per task over \(2\) independent runs and report pass@2 mean accuracy on ARC-1 evaluation set. 
More choices are analyzed in the appendix.

\noindent \textbf{(a) State update composition.}
As introduced in Section~\ref{subsec:update_modes}, 
we first compare three update variants: \textit{slot-only}, 
\textit{transport-only}, 
and \textit{complementary update}. 
To isolate this factor, all variants use only the per-task summary parameterization. Figure ~\ref{fig:update_ablation} shows a clear ordering; the complementary update performs best. 
This supports our hypothesis that ARC world-state transitions require both spatial transport and object-aware content updates, rather than pure routing alone.

\noindent \textbf{(b) Task-summary parameterization.}
We next compare three variants introduced in Section~\ref{subsec:task_summary}: 
\textit{shared-only}, 
\textit{per-task}, 
and \textit{hybrid}. 
To isolate this factor, all variants use the same complementary update mode. 
Figure~\ref{fig:summary_ablation} highlights an empirical trade-off. Shared-only summary
tokens provide the cleanest amortized formulation, but offer limited task-specific
capacity during test-time training. Adding per-task summary entries substantially
improves performance, suggesting that these entries act as adaptable placeholders for
task-specific rule information. The hybrid is slightly better than per-task-only,
indicating that shared summary queries still provide a useful amortized rule-induction
bias when combined with task-specific capacity.

\section{Limitations and Future Work}

\begin{figure}[h]
\vspace{-2mm}
\centering
\includegraphics[width=\linewidth]{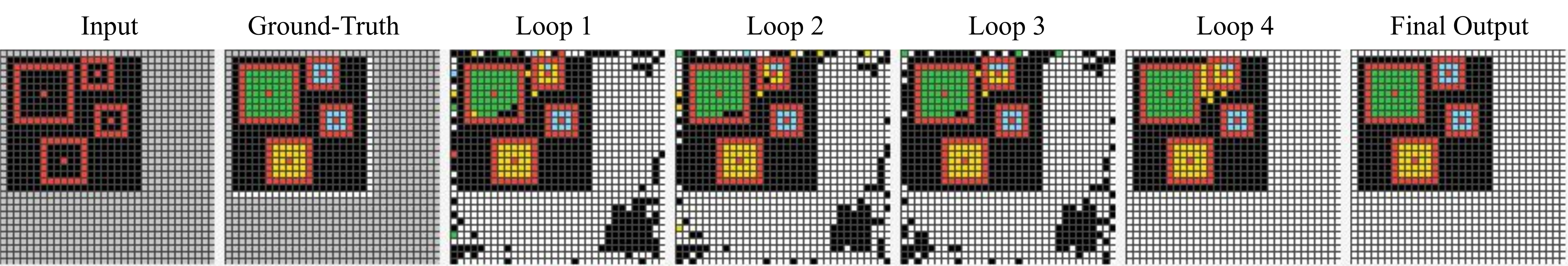}
\caption{Example rollout without stable semantics.}
\vspace{-3mm}
\label{fig:loop}
\end{figure}

Although Loop-OWM uses recurrence as a rollout mechanism, we do not observe stable,
human-interpretable semantics at individual loop steps, as shown in
Figure~\ref{fig:loop}. Ideally, different loops might correspond to reusable reasoning
stages such as localization, transport, recoloring, or boundary refinement. In practice,
however, computation is distributed across steps without a consistent cross-task pattern.
This is expected to some extent: 
standard ARC provides only input-output supervision and \textit{does not supervise intermediate states}; the internal rollout is therefore underdetermined as long as the final grid is correct.

Addressing this limitation requires trajectory-level supervision. As a first step, we
are building an ARC-2-derived dataset with principled intermediate-state annotations,
currently covering approximately 500 tasks. By supervising sub-transformations rather
than only final outputs, such data could make recurrent updates more semantically
grounded and move looped ARC models toward more inspectable world models.

\section{Conclusion}

We frame ARC as rule induction over visual-symbolic state transitions. Under this view, demonstrations define how a structured grid state should change, and solving a new query requires executing the inferred transition rather than merely predicting pixels.
Our proposed model instantiates this idea with object-aware grid interpretation, task-summary tokens, and recurrent transition updates that combine dense transport with
slot-conditioned writing. Results on ARC-1 and ARC-2, together with the targeted ablations,
show that this formulation provides an effective inductive bias for compact neural ARC
solvers. More broadly, the study suggests that ARC can serve not only as a benchmark for language- or program-based reasoning, but also as a controlled setting for learning
composable world-state dynamics.

\section*{Acknowledgments}

The authors thank their collaborators for their discussions and support throughout this work. This research was supported in part by a EuroHPC Joint Undertaking compute grant, which provided access to large-scale GPU resources and was essential for conducting the experiments in this paper.

\bibliography{bibliography_long,custom}

\appendix

\section{Appendix}
\label{sec:appendix}

\appendixtitlespacing

\subsection{Configurations and Model Size}
\label{app:param_count}

We report the training configurations in Table~\ref{tab:configurations}. 
In addition, Table~\ref{tab:param_breakdown} shows the parameter breakdown of our \(64/2\) model (\(64 \times 64\) canvas with patch size \(p=2\)).  
The model has \(10.63\)M parameters excluding the per-task summary table. 
This table is tied to the number of training task identifiers and therefore scales with the benchmark-specific task vocabulary rather than with the reusable model architecture. 
For ARC-2, the per-task summary table contains \(1000 \times (N^q_\mathrm{task}=)8 \times 256 \approx 2.048\)M parameters, where $N^q_\mathrm{task}$ defines the capacity of the summary query bank (see \ref{app:task_encoder_impl}).  
Most parameters are concentrated in three components: the task encoder, the transition core, and the decoder. Together, these account for approximately \(84.4\%\) of the model.
The slot module is shared between the task encoder and the looped predictor and contributes only \(10.6\%\) of the total parameter count. We use Muon~\cite{muon} optimizer with cosine lr-scheduler through training.  

\noindent \textbf{Number of recurrent steps.}
We ablate the number of recurrent rollout steps in the \(32/1\) configuration,
testing \(T \in \{3,4,5,6\}\). We ablate \(T \in \{3,4,5,6\}\) rollout steps and find that \(T=4\) gives the best validation performance, with no visible benefit from additional steps. We therefore use \(T=4\) throughout all experiments.

\begin{table}[h]
\centering
\small
\setlength{\tabcolsep}{0pt}
\renewcommand{\arraystretch}{1.08}

\begin{tabular}{@{}
p{0.45\linewidth}
@{\hspace{1em}\vrule width 0.4pt\hspace{0.1em}}
>{\raggedleft\arraybackslash}p{0.5\linewidth}
@{}}
\specialrule{\heavyrulewidth}{0pt}{0pt}
\rowcolor{gray!15}
\multicolumn{2}{@{}l@{}}{\textbf{\textit{offline training}}} \\

epochs                  & 200 \\
warmup epochs           & 0 \\
optimizer               & Muon, $\beta=0.95$ \\
batch size (per GPU)    & 64 \\ 
GPU type                & H200 $\times$4 \\
learning rate           & $1\mathrm{e}{-3}$ \\
learning rate scheduler & cosine (min $0.05\times$) \\
weight decay            & 0.01 \\
dropout                 & 0 \\
task query bank $N^q_\mathrm{task}$     & 8 \\
use loop-index embedding & true \\
loop steps              & 4 \\
$\mathcal{L}_\text{trans}$ weight & 0.1 \\

\specialrule{\lightrulewidth}{0pt}{0pt}
\rowcolor{gray!15}
\multicolumn{2}{@{}l@{}}{\textbf{\textit{test-time training}}} \\

epochs                  & 200 \\
warmup epochs           & 0 \\
optimizer               & Muon, $\beta=0.95$ \\
batch size              & 8 \\
learning rate           & $5\mathrm{e}{-4}$ \\
learning rate scheduler & cosine (min $0.05\times$) \\
weight decay            & 0.01 \\
dropout                 & 0 \\
task query bank $N^q_\mathrm{task}$     & 8 \\
use loop-index embedding & true \\
loop steps              & 4 \\
$\mathcal{L}_\text{trans}$ weight & 0.1 \\
views per test pair     & 30 \\
independent runs        & 5 \\

\bottomrule
\end{tabular}

\caption{\textbf{Training Configurations.}}
\label{tab:configurations}
\end{table}

\begin{table*}[t]
\centering
\small
\setlength{\tabcolsep}{3pt}
\renewcommand{\arraystretch}{1.12}

\begin{tabularx}{\textwidth}{@{}
>{\raggedright\arraybackslash}p{0.18\textwidth}
>{\raggedleft\arraybackslash}p{0.08\textwidth}
>{\raggedleft\arraybackslash}p{0.08\textwidth}
>{\raggedright\arraybackslash}p{0.15\textwidth}
>{\raggedright\arraybackslash}p{0.1\textwidth}
>{\raggedright\arraybackslash}X
@{}}
\toprule
Component & \#Params & Share & Type & Hidden dim & Description \\
\midrule
Task encoder $\psi_\theta(\cdot)$
& 3.17M & 29.8\% & $4$ blocks
& 256 
& Cross- and Self-attn, MLP (ff$=512$) \\

Transition $F_\theta(\cdot)$
& 3.16M & 29.8\% & $6$ blocks
& 256
& Looped transformer core, 4 steps \\

Decoder $\mathrm{Dec}_\theta(\cdot)$
& 2.63M & 24.8\% & Cross-Attn + MLP
& 256
& Slot mixer, renderer, output head (ff$=512$) \\

Slot encoder $S_\theta(\cdot)$
& 1.12M & 10.6\% & Slot attention
& $256$
& Shared color-prototype, 24 slots, 3 iters \\

Patch embedding
& 0.26M & 2.5\% & Conv2d ($1{\times}1$)
& 256
& Convolutional patch embedding \\

2D pos embedding
& 0.26M & 2.5\% & sin-cosine
& $1024 \times 256$
& Positional embedding table \\

Output head
& 0.01M & 0.1\% & Linear
& $256 \to 12$
& Color prediction head \\

Other small modules
& $<0.01$M & $<0.1\%$ & others
& 256
& Color-, role-, step-embeddings, LayerNorm \\

\midrule
Total
& 10.64M & 100\% &  &  & Excluding per-task summary bank \\
\bottomrule
\end{tabularx}

\caption{
Parameter breakdown of the Loop-OWM (64/2) configuration, excluding the per-task
summary table. The total model size is \(10.64\)M parameters. The task encoder,
transition core, and decoder dominate the parameter count, while the shared
slot tokenizer remains comparatively small.
}
\label{tab:param_breakdown}
\end{table*}
 
\subsection{Data Preprocessing}
Following the visual-canvas setup in VARC \citep{hu2025varc}, we place ARC grids, whose spatial size is at
most \(30 \times 30\), on an image-like canvas. We use a \(2\times\) scaling ratio with
a \(64 \times 64\) canvas and patch size \(\mathrm{p}=2\), producing a \(32 \times 32\) token
grid with \(L=1024\) patch tokens. This keeps the transformer sequence length comparable
to a \(32 \times 32\) canvas with \(1 \times 1\) patches, while allowing each
\(2 \times 2\) patch to contain richer local color patterns.

Since ARC input and output grids may have different raw shapes, we follow \citep{hu2025varc} and handle
shape prediction on the fixed canvas.
Canvas cells outside the placed grid receive an \(\texttt{[IGNORE]}\) token, while output grids additionally use a \(\texttt{[PAD]}\) token to mark the right column and bottom row immediately adjacent to the valid output region
\footnote{In our implementation, \texttt{[IGNORE]} and \texttt{[PAD]} correspond to \texttt{IGNORE\_INDEX}=10 and \texttt{PAD\_INDEX}=11, respectively.}. 
The grid cross-entropy is computed on all non-\(\texttt{[IGNORE]}\) cells; \(\texttt{[PAD]}\) tokens are supervised as meaningful boundary tokens, allowing the model to recover the output shape from its own predictions. 
At inference time, we locate the rightmost and bottommost predicted \(\texttt{[PAD]}\) tokens and crop the canvas accordingly. Following \citep{hu2025varc}, we also apply a foreground mask in the predictor self-attention layers by masking keys corresponding to \(\texttt{[IGNORE]}\) cells; \(\texttt{[PAD]}\) keys are left unmasked because they encode valid output-boundary information.

During training, we apply VARC-style canvas augmentations, including random translation,
scale changes, rotations, flips, and color permutations. Spatial augmentations are applied
consistently to the input and output grids of each pair so that the underlying task rule
is preserved.

\subsection{Positional Embeddings}
We use absolute 2D sine-cosine positional embeddings~\citep{he2021masked} at several sites in the model.
At the input, each cell token is formed by concatenating a color embedding slice with a 2D positional embedding, rather than adding position to color. This keeps color and position in separate channels at initialization, which preserves the initial dot-product alignment between color-prototype slot seeds and same-color patches. 
Importantly, this does not force slots to ignore position: $S_\theta(\cdot)$ applies learnable projections to the full token dimension, so the model can still learn to use positional information when useful.

For slot-mediated update (Sec.~\ref{subsec:update_modes}), 
we also add 2D positional embeddings to the per-patch queries $\mathbf{Q} = \{q_j\}_{j=1}^{L}$ in the slot mixer \(\mathcal{M}_\theta^{\mathrm{s2g}}(\cdot)\), enabling the allocator to route slot content to specific grid positions. 
In addition, the Transformer-based transition module $F_\theta(\cdot)$ applies 2D RoPE \citep{RoPE} to the image-token attention block, providing a relative positional signal complementary to the absolute
sine-cosine embeddings.

\begin{figure*}[t]
\centering
\includegraphics[width=\textwidth]{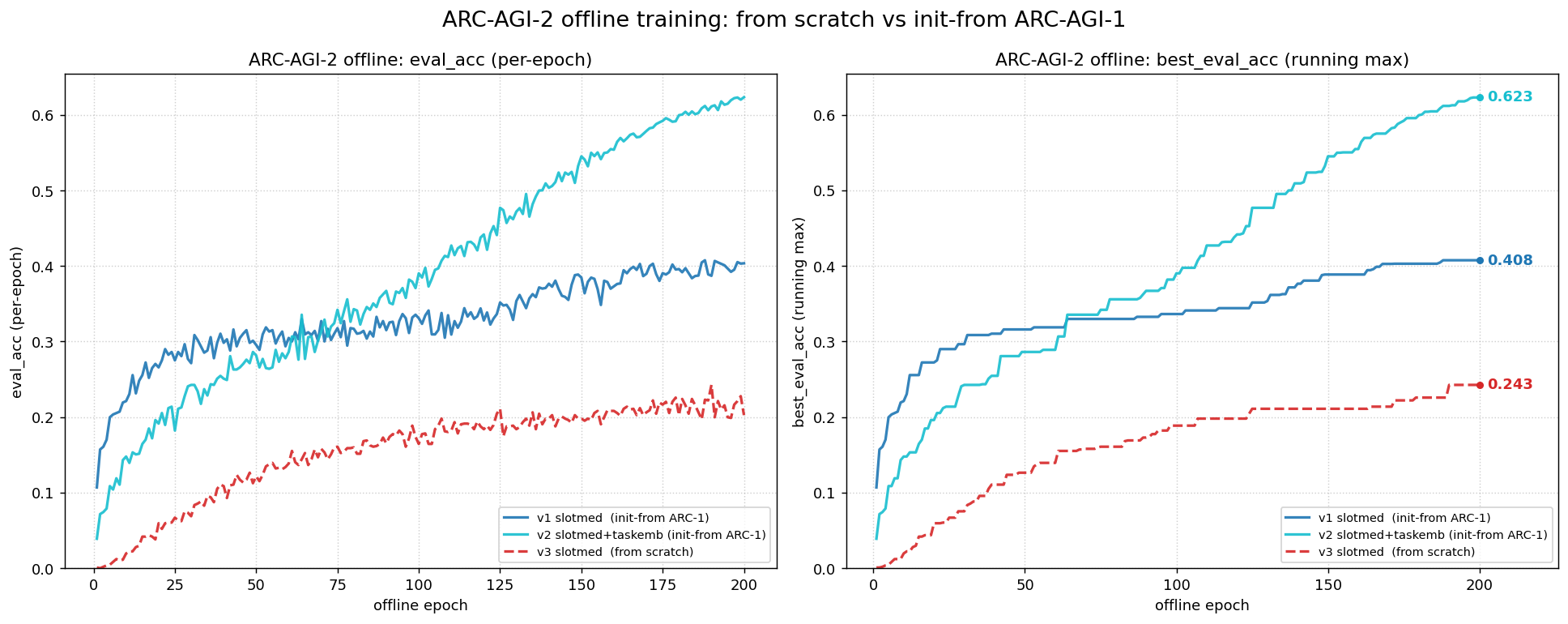}
\caption{
ARC-2 offline training from scratch versus initialization from an ARC-1
checkpoint trained for 200 offline epochs. 
Left: per-epoch validation accuracy. Right: running maximum validation accuracy. 
}
\label{fig:arc2_transfer}
\end{figure*}

\subsection{Task Encoder Implementation}
\label{app:task_encoder_impl}

The task encoder \(\psi_\theta(\cdot)\) (introduced in Section \ref{subsec:task_summary}) maps the
demonstration set \(\mathcal{D}\) to a sequence of context tokens
\(z_{\mathcal{D}}\) that condition the looped transition model $F_\theta(\cdot)$. 

For each demonstration pair, let
\(H_{\mathrm{in}}\) and \(H_{\mathrm{out}}\) denote the encoded input and output features. 
We first compute a demo-side transition
\(T_{\mathrm{demo}}\) as a softmax-normalized feature similarity from input units to
output units. This provides a lightweight alignment between the two grids and is used only to construct \textit{transition-aware features} for the task encoder. 
Using this alignment, we represent each demonstration pair with
\(r \in \{3,4\}\) role streams:
(1) the input features \(H_{\mathrm{in}}\), (2) the output features \(H_{\mathrm{out}}\), (3) the
post-transport residual
\[
\Delta_{\mathrm{demo}}
=
H_{\mathrm{out}} - T_{\mathrm{demo}} H_{\mathrm{in}},
\]
and, (4) optionally, a position-flow stream obtained by transporting the per-unit positional
embeddings through the same transition \(T_{\mathrm{demo}}\). The residual stream
highlights content that cannot be explained by directly transporting the input features,
while the optional position-flow stream exposes the spatial displacement implied by the
demo-side transition.

Each role stream is shifted by a learnable role embedding
\(e_\mathrm{role} \in \mathbb{R}^{d}\), shared across demonstrations and tasks. The role-tagged streams are then concatenated to each demo-pair and projected to the model dimension. This produces the demo-pair tokens attended by the task-summary query bank.

The summary query bank consists of $N^q_\mathrm{task}$($=8$) learnable queries. These queries cross-attend
to the demo-pair tokens through \(D_{\mathrm{task}}\) ($=4$) layers, producing the task context tokens \(z_{\mathcal{D}}\). Depending on the summary-token variant introduced in Section \ref{subsec:task_summary}, the initial queries may be shared across all tasks, retrieved from a per-task lookup table, or formed by combining shared and task-specific queries. The
resulting context tokens are concatenated with a loop-index embedding at each rollout step before conditioning the transition model \(F_\theta(\cdot)\).

\noindent \textbf{Role embeddings.}
Inside the task encoder, each demonstration pair is decomposed into \(r \in \{3,4\}\)
role-specific streams at a common unit granularity: slots in slot-encoder mode and
patches otherwise. The streams correspond to the demonstration input, demonstration
output, post-transport residual, and optionally a position-flow stream obtained by
transporting per-unit positional embeddings through the same demo-side transition. We
add a learnable role embedding \(e_r \in \mathbb{R}^{d}\) to each stream, shared across
demonstrations and units. The role-tagged streams are then concatenated along the channel
dimension and linearly projected to the model dimension, producing the demo-pair tokens
attended by the task-summary queries. The additive role tag complements channel
concatenation by giving each stream a learnable role anchor before projection.

\subsection{Cross-benchmark Transfer Learning}
\label{app:cross_benchmark_transfer}

ARC-1 and ARC-2 share the same grid and color interface, but ARC-2 defines a new and more challenging distribution of tasks. This creates a natural transfer
setting for evaluating training efficiency:  
if the model has learned general visual-symbolic machinery on ARC-1, then adapting to ARC-2 should require less training than learning the same machinery from scratch. 

We study 1-to-2 transfer by first training Loop-OWM on ARC-1 for 200 offline epochs, and then using the resulting checkpoint to initialize ARC-2 offline training. 
We copy all task-independent components; parameters tied to benchmark-specific task identities, such as per-task summary tables, are re-initialized for ARC--2. 
We reset the optimizer before training on ARC-2 and use cosine learning-rate scheduler for both runs.

Figure~\ref{fig:arc2_transfer} compares ARC-2 offline validation accuracy between model training from scratch (v3, \textcolor{myred1}{red curve}) with model initialized from ARC-1 checkpoints (v1, \textcolor{myblue1}{deeper-blue curve}). 
The transferred model converges faster and reaches substantially higher offline validation accuracy, indicating that ARC-1 pretraining provides a useful initialization for object-centric abstraction and task-conditioning mechanism. 
Since task-id-dependent parameters are not reused (the [+taskembed] curve (v2, \textcolor{myblue2}{lighter-blue}) isolates the effect of adding a per-task summary, which is re-initialized for ARC-2), this improvement between v1 and v3 cannot be explained by carrying over task-specific lookup memory for repeated ARC-1 tasks.




\end{document}